\documentclass[a4wide]{article}

\usepackage{fullpage}
\usepackage{times}
\usepackage{helvet}
\usepackage{courier}
\usepackage{listings}
\usepackage{latexsym}
\usepackage{amsmath}
\usepackage{amssymb}
\usepackage{graphicx}
\usepackage{url}

\newcommand{\ail}{\textsc{AIL}}
\newcommand{\ajpf}{\textsc{AJPF}}
\newcommand{\prism}{\textsc{Prism}}
\newcommand{\jpf}{JPF}
\newcommand{\java}{Java}
\newcommand{\mcapl}{MCAPL}

\newcommand{\imp}{\Rightarrow}
\newcommand{\always}{\ensuremath{\Box}}

\newlength{\myfigures}

\newcounter{lst}

\lstdefinestyle{deglisting}{basicstyle=\footnotesize\sffamily, mathescape=true, frame=tb,  numbers=right, numberstyle=\footnotesize, stepnumber=1, numbersep=-5pt, captionpos=b}

\lstdefinestyle{listingl}{basicstyle=\footnotesize\sffamily, mathescape=true, numberstyle=\footnotesize, stepnumber=1, numbersep=-5pt, captionpos=b}

\lstdefinestyle{deg}{basicstyle=\sffamily, mathescape=true}

\lstnewenvironment{listing}[3]{
  \noindent        
  \refstepcounter{lst}         
  \label{code:#1}  
\begin{tabular}{p{.94\columnwidth}} \\ \hline  {\normalsize \textbf{Code Listing \arabic{lst}} #2} \\  \end{tabular} 
\lstset{language=#3,          
  basicstyle=\footnotesize\sffamily,
  xleftmargin=10pt,    
  mathescape=true,    
  frame=tb,    
  numbers=right,    
  numberstyle=\footnotesize,     
  stepnumber=1,     
  numbersep=-5pt}}{}

\lstdefinelanguage{DEG}{%
    morekeywords={Outcome,Scores,Ethical,Precedence},
    morecomment=[l]{//},
 literate= {>}{{$>$}}1
           {=}{{$=$}}1
}

\title{Towards Verifiably Ethical Robot Behaviour} 
\author{Louise A. Dennis $\And$ Michael Fisher \\ 
  Department of Computer Science\\ University of Liverpool, UK\\ 
  \texttt{\{L.A.Dennis,MFisher\}@liverpool.ac.uk}\\ 
  $\And$
   Alan F. T. Winfield\\ 
   Bristol Robotics Laboratory \\ University of the West of England, UK\\
        \texttt{Alan.Winfield@uwe.ac.uk}}

\begin{document}
\maketitle
\setlength{\myfigures}{0.4\textwidth}

\begin{abstract}Ensuring that autonomous systems work \emph{ethically}
  is both complex and difficult. However, the idea of having an
  additional `governor' that assesses options the system has, and
  prunes them to select the most ethical choices is well
  understood. Recent work has produced such a governor consisting of
  a `consequence engine' that assesses the likely future
  outcomes of actions then applies a Safety/Ethical logic to select actions. Although this is appealing, it is
  impossible to be certain that the most ethical options are actually
  taken. In this paper we extend and apply a well-known agent
  verification approach to our consequence engine, allowing us to verify
  the correctness of its ethical decision-making.
\end{abstract}

\section{Introduction}
It is widely recognised that autonomous systems will need to conform
to legal, practical and \emph{ethical} specifications. For instance,
during normal operation, we expect such systems to fulfill their goals
within a prescribed legal or professional framework of rules and
protocols.  However, in exceptional circumstances, the autonomous
system may choose to ignore its basic goals or break legal or
professional rules in order to act in an ethical fashion, e.g., to
save a human life. But, we need to be sure that the system will only
break rules for justifiably ethical reasons and so we are here
concerned with the verification of autonomous systems and, more
broadly, with the development of \emph{verifiable autonomous systems}.

This paper considers a technique for developing verifiable ethical
components for autonomous systems, and we specifically consider
the \emph{consequence engine} proposed by~\cite{Winfield13}. This
consequence engine is a discrete component of an autonomous
system that integrates together with methods for action selection in
the robotic controller. It evaluates the outcomes of actions using
simulation and prediction, and selects the most ethical option using a
\emph{safety/ethical logic}.  In Winfield et al.~\cite{Winfield13}, an example of such
a system is implemented using a high-level Python program to control
the operation of an e-puck robot~\cite{mondada09} tracked with a
VICON system. This approach tightly couples the ethical reasoning with
the use of standard criteria for action selection and the
implementation was validated using empirical testing.

In addition, given the move towards configurable, component-based
plug-and-play platforms for robotics and autonomous systems,
e.g.~\cite{verfaillie06,dennis14,288}, we are interested in decoupling
the ethical reasoning from the rest of the robot control so it appears
as a distinct component. We would like to do this in a way that allows
the consequence engine to be verifiable in a straightforward
manner.

This paper describes the first steps towards this.  It develops a
declarative language for specifying such consequence engines as agents
implemented within the \emph{agent infrastructure layer toolkit}
(\ail). Systems developed using the \ail\ are verifiable in the
\ajpf\ model-checker~\cite{MCAPL_journal} and can integrate with
external systems such as MatLab simulations~\cite{IEEE13}, and Robotic
Operating System (ROS) nodes~\cite{ros-ail-TR}. Having developed the
language, we then reimplement a version of the case study reported
in Winfield et al.~\cite{Winfield13} as an agent and show how the operation of the
consequence engine can be verified in the \ajpf\ model checker. We
also use recently developed techniques to show how further investigations of the
system behaviour can be undertaken by exporting a model from \ajpf\ to
the \prism\ probabilistic model checker.

\section{Background}
\subsection{An Internal Model Based Architecture}
\begin{figure}
\begin{center}
\includegraphics[width=0.8\textwidth]{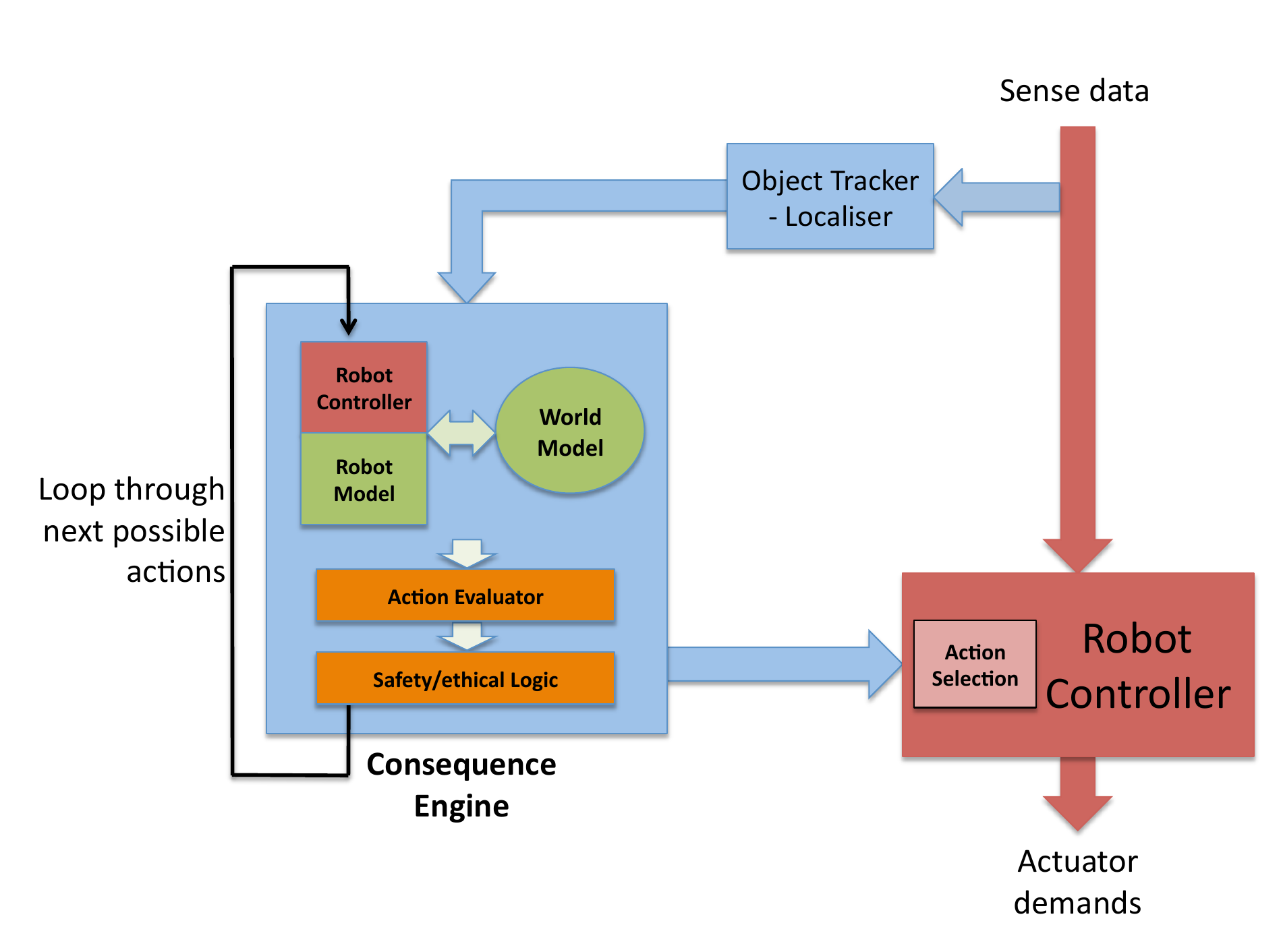}
\end{center}
\caption{Internal-model based architecture.  Robot control data flows are shown in red (darker shaded); the Internal Model data flows in blue (lighter shaded).}
\label{fig:architecture}
\end{figure}

Winfield et al.~\cite{Winfield13} describe both the abstract
architecture and concrete implementation of a robot that contains a
consequence engine. The engine utilises an internal model of the robot
itself and its environment in order to predict the outcomes of actions
and make ethical and safety choices based on those predictions.  The architecture
for this system is shown in Figure~\ref{fig:architecture}.
In this architecture, the consequence engine intercepts the robot's
own action selection mechanism. It runs a simulation of all available
actions and notes the outcomes of the simulation.  These outcomes are
evaluated and selected using a \emph{Safety/Ethical Logic} layer (SEL).

Winfield et al.~\cite{Winfield13} consider a simple scenario in
which a human is approaching a hole. In normal operation the robot
should select actions which avoid colliding with the human but, if the
robot's inference suggests the human will fall in the hole then it may
opt to collide with the human. While this is ``against the rules'', it
is a more ethical option as it avoids the greater harm of the human falling into the hole. In
order to do this, the paper suggests scoring the outcomes of the
actions for each of the actors (the human and the robot) -- e.g., 0 if
the actor is safe, 4 if the actor is involved in a collision and 10 if
the actor falls in the hole.  It then recommends a simple if-then
logic for selecting actions based on these values.

\begin{lstlisting}[basicstyle=\footnotesize\sffamily]
IF for all robot actions, 
            the human is equally safe
THEN (* default safe actions *) 
  output safe actions
ELSE (* ethical action *)
  output action(s) for 
            least unsafe human outcome(s)
\end{lstlisting}

\noindent A version of this architecture was implemented on e-pucks
(small, relatively simple, robots). The basic activity is for the
robot to attempt to reach some target location while avoiding a hole
in the ground. Two humans (actually simulated by additional e-pucks in
this experiment) were also programmed to move towards the hole and then
the robot could choose to move towards these in an attempt to get them
to divert using their own inbuilt avoidance mechanisms in order to
prevent them falling into the hole. A number of experiments were
carried out.  In all situations the robot avoided falling into the
hole itself. When there was a single other robot (representing the
human that is in danger) it successfully managed to divert the
``human'' on all occasions.  When a third robot (representing a second
human) was introduced into the problem, the robot rescued at least one
``human'' in about 58\% of runs and rescued both in 9\% of runs. These
outcomes depended upon both noise and the starting conditions effecting
which additional robot moved first and whether the e-puck had time to
reach both of them.

The actual implementation of the basic ethical action selection in this system
was based on \emph{potential functions}. Each action was
assigned a score based upon the weighted sum of how close it took the
e-puck to the goal position, whether the e-puck was placed in danger,
whether one of the other ``humans'' was placed in danger, and how
close the action would take the e-puck to the ``humans''.  The
system then simply selected the action with the highest score.

\subsection{Verifying Autonomous Systems using AJPF}
Formal verification is essentially the process of assessing whether a
specification, given in formal logic, is true of the system in
question. For a specific logical property, $\varphi$, there are many
different approaches to achieving
this~\cite{Fetzer88:veryidea,dMLP79,BoyerMoore81}, ranging from
deductive verification against a logical description of the system
$\psi_S$ (i.e., $\vdash \psi_S\imp \varphi$) to the algorithmic
verification of the property against a formal model of the system, $M$
(i.e., $M\models\varphi$). The latter has been extremely successful in
Computer Science and Artificial Intelligence, primarily through the
\emph{model checking} approach~\cite{Clarke00:MC}. This takes an
executable model of the system in question, defining all the system's
possible executions, and then checks a logical property against this
model (and, hence, against all possible executions).

Whereas model checking involves assessing a logical specification
against all executions of a \emph{model} of the system, an alternative
approach is to check a logical property directly against all
\emph{actual} executions of the system. This is termed the \emph{model
  checking of programs}~\cite{VisserHBPL03} and crucially depends on
being able to determine all executions of the actual program. In the
case of \java{}, this is feasible since a modified virtual machine can
be used to manipulate the program executions. The Java Pathfinder
(\jpf) system carries out formal verification of \java{} programs in
this way by analysing all the possible execution
paths~\cite{VisserHBPL03}. This avoids the need for an extra level of
modelling and ensures that the verification results truly apply to the
real system.

In the examples discussed later in this paper we use the \mcapl{}
framework which includes a model checker for agent programs built on
top of \jpf. As this framework is described in detail
in~\cite{MCAPL_journal}, we only provide a brief overview
here. \mcapl{} has two main sub-components: the \ail-toolkit for
implementing interpreters for belief-desire-intention (BDI) agent
programming languages and the \ajpf\ model checker.

Interpreters for BDI languages are programmed by instantiating the
\java-based {\em AIL toolkit}~\cite{AIL07:ProMAS}. Here, an agent
system can be programmed in the normal way for the programming language
but then runs in the AIL interpreter which in turn runs on top of the
Java Pathfinder (\jpf) virtual machine.

Agent \jpf\ (\ajpf) is a customisation of \jpf\ that is specifically
optimised for AIL-based language interpreters.  Agents programmed in
languages that are implemented using the \ail{}-toolkit can thus be
formally verified via \ajpf. Furthermore if they run in an environment
programmed in \java, then the whole agent-environment system can be
model checked.  Common to all language interpreters implemented using
the \ail\ are the \ail-agent data structures for \emph{beliefs},
\emph{intentions}, \emph{goals}, etc., which are subsequently accessed
by the model checker and on which the logical modalities of a property
specification language are defined.

The system described in Winfield et al.~\cite{Winfield13} is not explicitly a BDI
system or even an agent system, yet it is based on the concept of a software system that forms
some component in a wider environment and there was a moderately
clear, if informal, semantics describing its operation, both of which are key assumptions underlying the \mcapl{} framework. We therefore
targeted \ajpf\ as a preliminary tool for exploring how such a
consequence engine might be built in a verifiable fashion, especially
as simple decision-making within the safety/ethical logic could be
straightforwardly captured within an agent.

\section{Modelling a Consequence Engine for AJPF}

Since \ajpf\ is specifically designed to model check systems
implemented using \java\ it was necessary to re-implement the
consequence engine and case study described in Winfield et al.~\cite{Winfield13}.

We implemented a \emph{declarative consequence engine} in the \ail\ as
a simple language governed by two operational semantic rules, called
\emph{Model Applicable Actions} and \emph{Evaluating
  Outcomes}. Semantically, a consequence engine is represented as a
tuple $\langle ce, ag, \xi, A, An, SA, EP, f_{ES} \rangle$ where:
\begin{itemize}
\item $ce$ and $ag$ are the names of the consequence engine and the agent it is
linked to;
\item $\xi$ is an external environment (either the real world, a
simulation or a combination of the two);
\item $A$ is a list of $ag$'s actions that are currently applicable;
\item $An$ is a list of such actions annotated with outcomes;
\item $SA$ is a filtered list of the applicable actions, indicating
  the ones the engine believes to be the most ethical in the current
  situation;
\item $EP$ is a precedence order over the actors in the environment
  dictating which one gets priority in terms of ethical outcomes; and
\item $f_{ES}$ is a map from outcomes to an integer indicating their
  ethical severity.
\end{itemize}

\begin{equation}
\frac{An' = \{\langle a, os\rangle \mid a \in A \wedge os = \xi.model(a)\}}
{\langle ce, ag, \xi, A, An, SA, EP, f_{ES}\rangle \rightarrow 
\langle ce, ag, \xi, A, An', SA, EP, f_{ES}\rangle
}
\label{eq:modelaa}
\end{equation}

\noindent The operational rule for \emph{Model Applicable Actions} is
shown in \eqref{eq:modelaa}.  This invokes some model or simulation in
the environment ($\xi.model(a)$) that simulates the effects of $ag$
taking each applicable action $a$ and returns these as a list of
tuples, $os$, indicating the outcome for each actor, e.g., $\langle
human, hole\rangle$ to indicate that a human has fallen into a hole.
The consequence engine replaces its set of annotated actions with this
new information.

\begin{equation}
\frac{SA' = f_{ep}(EP, An, f_{ES}, A)}
{\langle ce, ag, \xi, A, An, SA, EP, f_{ES}\rangle \rightarrow 
\langle ce, ag, \xi, A, An, SA', EP, f_{ES}\rangle}
\label{eq:ethicalo}
\end{equation}

\noindent The operational rule for \emph{Evaluating Outcomes},
specifically of the ethical actions, is shown in \eqref{eq:ethicalo}.
It uses the function $f_{ep}$ to select a subset of the agent's
applicable actions using the annotated actions, the precedence order
and an evaluation map as follows:

\begin{equation}
f_{ep}([], An, f_{ES}, SA) = SA 
\end{equation}
\begin{equation}
f_{ep}(h::T, An, f_{ES}, SA) =  f_{ep}(T, An, f_{ES}, f_{me}(h, An, f_{ES}, SA)) 
\end{equation}
$f_{ep}$ recurses over a precedence list of actors (where $[]$ indicates the empty list and $h::T$ is element $h$ in front of a list $T$).  It's purpose is to filter the set of actions down just to those that are best, ethically, for the first actor in the list (i.e., the one whose well-being has the highest priority) and then further filter the actions for the next actor in the list and so on.  The filtering of actions for each individual actor is performed by $f_{me}$.

\begin{equation}
f_{me}(h, An, f_{ES}, A ) =  
\quad \{a \mid a \in A \wedge 
  \forall a'\neq a \in A. \sum_{\langle a, \langle h, out \rangle \rangle \in An} f_{ES}(out) \leq \sum_{\langle a', \langle  h, out'\rangle \rangle \in An}f_{ES}(out')\}
\end{equation}
$f_{me}$ sums the outcomes for actor, $h$ given some action $a \in
A$ and returns the set of those where the sum has the lowest value.
For instance if all actions are safe for actor $h$ we can assume
that $f_{ES}$ maps them all to some equal (low) value (say 0) and so
$f_{me}$ will return all actions.  If some are unsafe for $h$ then
$f_{ES}$ will map them to a higher value (say 4) and these will be
excluded from the return set.  

We sum over the outcomes for a given actor because either there may be
multiple unethical outcomes and we may wish to account for all of
them, or there may be multiple actors of a given type in the
precedence order (e.g., several humans) and we want to minimize the
number of people harmed by the robot's actions.

It should be noted that this implementation of a consequence engine is
closer in nature to the abstract description from Winfield et al.~\cite{Winfield13}
than to the implementation where \emph{potential functions} are used
to evaluate and order the outcomes of actions.  This allows certain
actions to be vetoed simply because they are bad for some agent high
in the precedence order even if they have very good outcomes lower in
the order.  However, this behaviour can be also be reproduced by
choosing suitable weights for the sum of the potential functions (and,
indeed, this is what was done in the implementation
in~\cite{Winfield13}).

It should also be noted (as hinted in our discussion of $f_{me}$) that
we assume a precedence order over types of agents, rather than
individual agents and that our model outputs outcomes for types of
agents rather than individuals. In our case study we consider only
outcomes for humans and robots rather than distinguishing between the
two humans. Importantly, nothing in the implementation prevents an
individual being treated as a type that contains only one object.
\begin{figure*}
\begin{center}
\includegraphics[width=.8\textwidth]{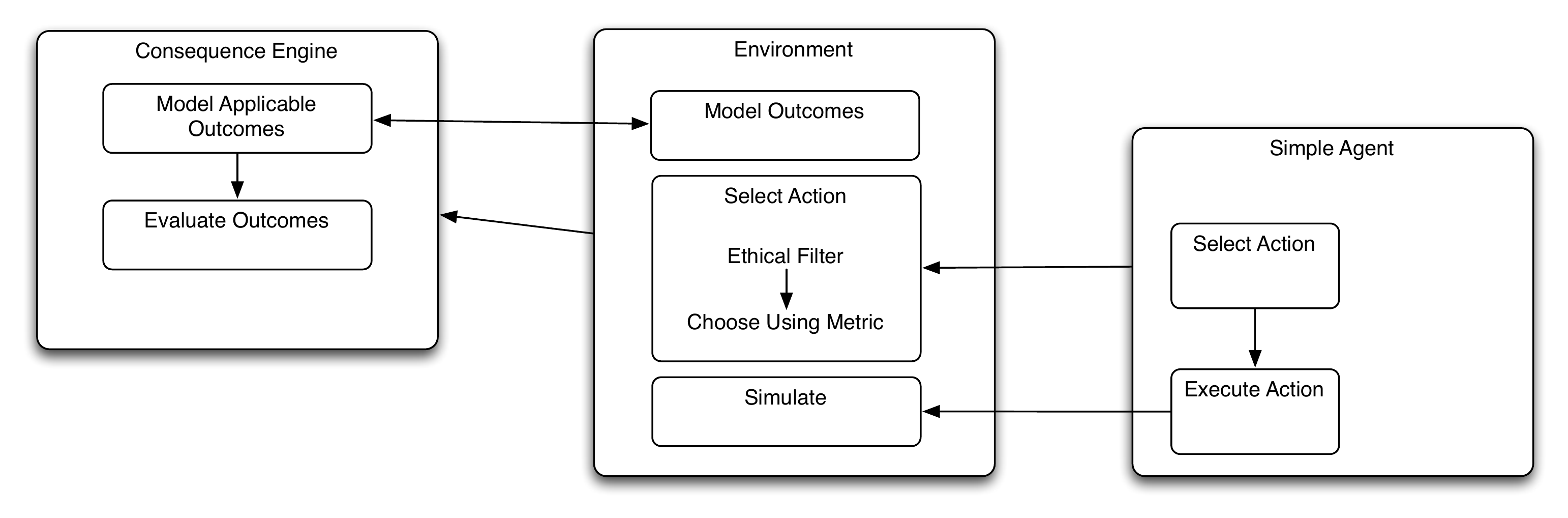}
\end{center}
\caption{Architecture for testing the AIL Version of the Consequence Engine}
\label{fig:ajpf_arch}
\end{figure*}

Our consequence engine language can be used to filter a set of actions
in any environment that can provide a suitable modelling capability.

\paragraph{Implementing a Robot}  In order to test the operation of 
consequence engines such as this, we also created a very simple agent
language in which agents can have beliefs, a single goal and a number
of actions. Each agent invokes an external $\mathit{selectAction}$
function to pick an action from the set of those applicable (given the
agent's beliefs).  Once the goal is achieved then the agent stops.  In
our case we embedded the consequence engine within the call to
$\mathit{selectAction}$. First, the consequence engine would filter
the available actions down to those it considered most ethical and
then \emph{selectAction} would use a metric (in this example,
distance) to choose the action which would bring the agent closest to
its goal.

This simple architecture is shown in Figure~\ref{fig:ajpf_arch}.
Here, arrows are used to indicate flow of control. In the simple agent
first an action is selected and then it is executed. Selecting this
action invokes the $\mathit{selectAction}$ method in the environment
which invokes first the consequence engine and then a metric-based
selection before returning an action to the agent. (The two rules in
the consequence engine are shown.)  Execution of the action by the
agent also invokes the environment which computes the appropriate
changes to the agents' perceptions.

Note that our implementation of the consequence engine is independent
of this particular architecture. In fact it would desirable to have
the consequence engine as a sub-component of some agent rather than as
a separate entity interacting with the environment, as is the case in Winfield et al.~\cite{Winfield13}. 
However this
simple architecture allowed for quick and easy prototyping of our
ideas~\footnote{Indeed the entire prototype system took less than a
  week to produce.}

\section{Reproducing the Case Study}
\begin{figure}
\begin{center}
\includegraphics[width=0.4\columnwidth]{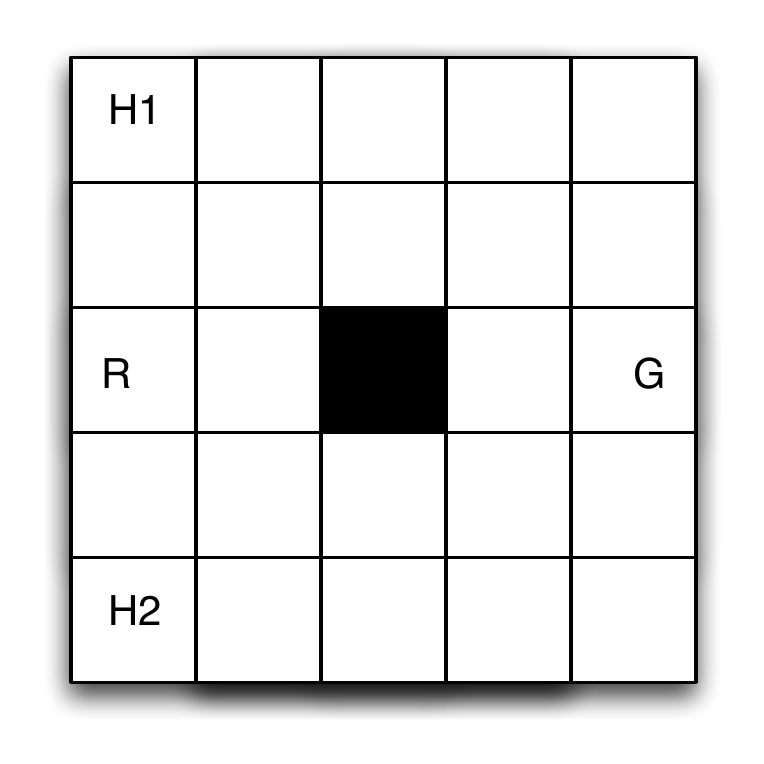}
\end{center}
\caption{Initial State of the Case Study Environment}
\label{fig:initial}
\end{figure}

We reproduced the case study described in Winfield et al.~\cite{Winfield13}.  Since
all parts of the system involved in the verification needed to exist
as Java code, we created a very simple simulated environment
consisting of a 5x5 grid.  Note that we could not reproduce the case study with full fidelity since we required a finite state space and the original case study took place in the potentially infinite state space of the physical world. The grid had a hole in its centre and
a robot and two humans represented in a column along one side.  At
each time step the robot could move to any square while there was a
50\% chance that each of the humans would move towards the hole.  The
initial state is shown in Figure~\ref{fig:initial}.  The robot,
\textsf{R}, can not reach the goal, G, in a single move and so will
move to one side or the other.  At the same time the humans,
\textsf{H1} and \textsf{H2}, may move towards the hole (central
square).

The agent representing the consequence engine is shown in code
listing~\ref{code:deg}.  Lines 6-7 define the map of outcomes to
values $f_{ES}$ and line 12 gives the precedence ordering.

\begin{listing}{deg}{Ethical Governor}{DEG}
:name: ethical_g
:agent: robot

:Outcome Scores:

safe = 0
collision = 4
hole = 10

:Ethical Precedence:

human > robot
\end{listing}

\noindent The actions available to the simple agent were all of the
form \lstinline{moveTo(X, Y)} where \lstinline{X} and \lstinline{Y}
were coordinates on the grid. A Breseham based super-cover line
algorithm~\cite{bresenham} was used to calculate all the grid squares
that would be traversed between the robot's current position and the
new one.  If these included either the hole or one of the ``humans''
then the outcomes $\langle robot, hole \rangle$ and $\langle robot,
collision \rangle$ together with $\langle human, collision \rangle$
were generated as appropriate.  If either of the ``humans'' occupied
a square adjacent to the hole then the outcome $\langle human, hole
\rangle$ was also generated.

\subsection{Results}
\begin{figure}
\begin{center}
\includegraphics[width=0.4\columnwidth]{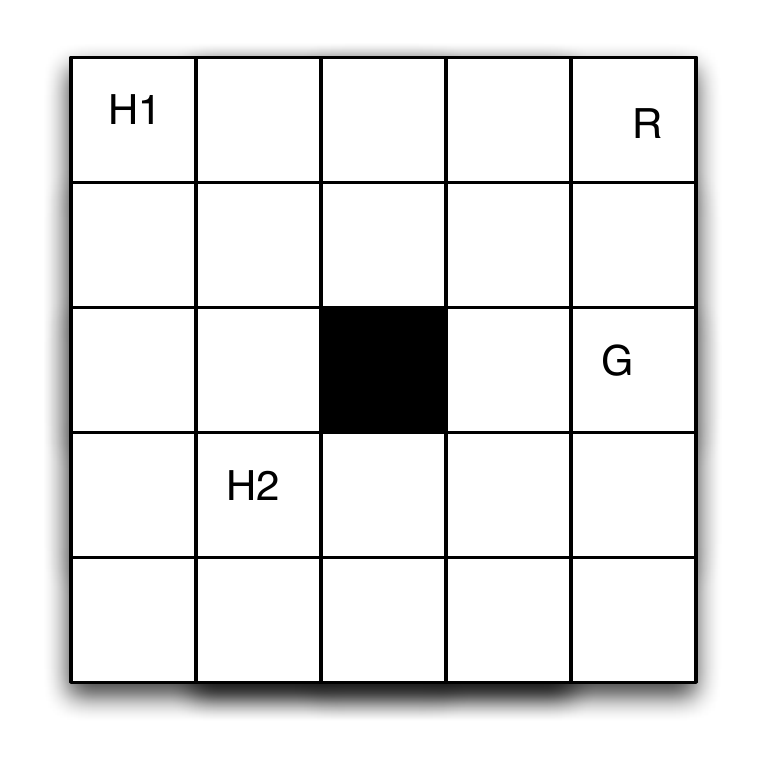}
\end{center}
\caption{Situation where the Robot can not save the Human}
\label{fig:hole}
\end{figure}

We were able to model check the combined program in \ajpf\ and so
formally verify that the agent always reached its target. However, we
were not able to verify that the ``humans'' never fell into the hole
because in several situations the hole came between the agent and the
human.  One such situation is shown in Figure~\ref{fig:hole}. Here,
Human \textsf{H2} will fall into the hole when it takes its next step
but the robot \textsf{R} cannot reach it in a single straight line
without itself falling into the hole before it reaches the human.

Since we were particularly interested in verifying the performance of
the consequence engine we adapted the modelling method in the
environment to assert \emph{percepts} (declarative statements the
robot could perceive) whenever one of the humans was in danger and
whenever there was a safe path for the robot to reach a human.  These
percepts had no effect on the execution of the program but their
existence could be checked by \ajpf's property specification language.
Using these percepts we were able to verify \eqref{eq:prop} where
$\always$ is the linear temporal operator meaning ``always'' and $B(r,
p)$ means that ``robot $r$ believes $p$ to be true''.
So~\eqref{eq:prop} reads that it is always the case that if the robot
believes $h1$ is in danger and it can find a safe path to $h1$ then it
will always be the case that the robot never believes $h1$ has fallen
in the hole.  We also proved the equivalent property for $h2$.

\begin{figure*}[bht]
\begin{equation}
\always (B(r, danger(h1)) \wedge B(r, path\_to(h1))) \rightarrow \always \neg B(r, h1(hole)) 
\label{eq:prop}
\end{equation}
\begin{multline}
\always (B(ce, sel(a_1)) \wedge B(ce, outcome(a_1, human(hole))))
\rightarrow  \\ 
B(ce, outcome(a_2, human(hole))) \wedge 
B(ce, outcome(a_3, human(hole))) \wedge 
B(ce, outcome(a_4, human(hole)))\label{eq:prop2}
\end{multline}
\begin{multline}
\always (B(ce, sel(a_1)) \wedge B(ce, outcome(a_1, robot(hole))))
\rightarrow \\ ( B(ce, outcome(a_2, human(hole))) \vee B(ce, outcome(a_2, robot(hole))) \vee B(ce, outcome(a_2, human(collision)))  )
\wedge \\ ( B(ce, outcome(a_3, human(hole))) \vee B(ce, outcome(a_3, robot(hole))) \vee B(ce, outcome(a_3, human(collision))) ) \wedge \\ ( B(ce, outcome(a_4, human(hole))) \vee B(ce, outcome(a_4, robot(hole))) \vee B(ce, outcome(a_4, human(collision))) ) \label{eq:prop3}
\end{multline}

\end{figure*}

It should be noted that we would not necessarily expect both the above to be the case because, in the situation where both H1 and H2 move simultaneously towards the hole, the robot would have to choose which to rescue and leave one at risk.  In reality it turned out that whenever this occurred the hole was between the robot and human 2 (as in figure~\ref{fig:hole}).  This was an artifact of the fact that the humans had to make at least one move before the robot could tell they were in danger.  The robot's first move was always to the far corner since this represented a point on the grid closest to the goal that the robot could safely reach.  The outcome would have been different if action selection had been set up to pick at random from all the points the robot could safely reach that were equidistant from the hole.

We were also able to export our program model to the probabilistic
\prism\ model checker, as described in~\cite{DennisFW13}, in order to
obtain probabilistic results.  These tell us that human 1 never falls
in the hole while human 2 falls in the hole with a probability of
0.71875 (71.8\% of the time).  The high chance of human 2 falling in
the hole is caused by the robot's behaviour, moving into the far
corner, as described above.  These probabilities are very different
from those reported in Winfield et al's experimental set up.  This
is primarily because the environment used here is considerably cruder,
with the robot able to reach any point in the grid in a single time
step.  The behaviour of the humans is also different to that
implemented in~\cite{Winfield13} where the H robots proceeded
steadily towards the hole and the differences in behaviour were
determined by small variations in the precise start up time and direction of each
robot.

\section{Verifying the Consequence Engine in Isolation}
Following the methodology from~\cite{ASEpaper} we also investigated verifying the consequence engine in isolation without any specific environment.  To do this we had to extend the implementation of our declarative language to allow the consequence engine to have mental states which could be examined by \ajpf's property specification language.  In particular we extended the operational semantics so that information about the outcomes of all actions were stored as beliefs in the consequence engine, and so that the final set of selected actions were also stored as beliefs in the consequence engine.  We were then able to prove theorems about these beliefs.

We developed a special verification environment for the engine.  This
environment called the engine to select from four abstract actions,
$a_1$, $a_2$, $a_3$ and $a_4$.  When the consequence engine invoked
the environment to model the outcomes of these four actions then four
possible outcomes were returned $\langle human, hole \rangle$,
$\langle robot, hole\rangle$, $\langle human, collision\rangle$ and
$\langle robot, collision \rangle$.  Each of these four outcomes was
chosen independently and at random --- i.e., the action was returned
with a random selection of outcomes attached.  \ajpf\ then explored
all possible combinations of the four outcomes for each of the four
actions.

\subsection{Results}

Model-checking the consequence engine in listing~\ref{code:deg} with
the addition of beliefs and placed in in this new environment we were
able to prove (\ref{eq:prop2}): it is always the case that if $a_1$ is
a selected action and its outcome is predicted to be that the human
has fallen in the hole, then all the other actions are also predicted
to result in the human in the hole --- i.e., all other actions are
equally bad.

We could prove similar theorems for the other outcomes,
e.g. (\ref{eq:prop3})
%
which states that if $a_1$ is the selected action and it results in
the robot falling in the hole, then the other actions either result in
the human in the hole, the robot in the hole, or the human colliding
with something.

In this way we could verify that the consequence engine indeed captured the order of priorities that we intended.

\section{Related Work}
The idea of a distinct entity, be it software or hardware, that can be
attached to an existing autonomous machine in order to constrain its
behaviour is very appealing. Particularly so if the constraints ensure
that the machine conforms to recognised \emph{ethical}
principles. Arkin~\cite{ArkinUW:2012} introduced this idea of an
\emph{ethical governor} to autonomous system, using it to evaluate the
``ethical appropriateness'' of a plan of the system prior to its
execution. The ethical governor prohibits plans it finds to be in
violation with some prescribed ethical constraint.

Also of relevance Anderson and Anderson's approach, where
\emph{machine learning} is used to `discover' ethical principles,
which then guide the system's behaviour, as exhibited by their
humanoid robot that ``takes ethical concerns into consideration when
reminding a patient when to take medication''~\cite{Anderson2:08}. A
range of other work, for example in~\cite{AndersonA11,Powers:2006},
also aims to construct software entities (`agents') able to form
ethical rules of behaviour and solve ethical dilemmas based on
these. The work of~\cite{WiegelB09} provides a logical framework for
\emph{moral reasoning}, though it is not clear whether this is used to
modify practical system behaviour.

Work by one of the authors of this paper (Winfield) has involved
developing and extending a generalised methodology for safe and
ethical robotic interaction, comprising both \emph{physical} and
\emph{ethical} behaviours. To address the former, a \emph{safety
  protection system}, serves as a high-level safety enforcer by
governing the actions of the robot and preventing it from performing
unsafe operations~\cite{Woodman12}. To address the latter, the
\emph{ethical consequence engine} studied here has been
developed~\cite{Winfield13}.

There has been little direct work on the formal verification of
ethical principles in practical autonomous systems. Work of the first
two authors has considered the formal verification of ethical
principles in autonomous systems, in particular autonomous
vehicles~\cite{dennis14ethical}. In that paper, we propose and
implement a framework for constraining the plan selection of the
rational agent controlling the autonomous vehicle with respect to
ethical principles.  We then formally verify the ethical extent of the
agent, proving that the agent never executes a plan that it knows is
`unethical', unless it does not have any ethical plan to choose. If
all plan options are such that some ethical principles are violated,
it was also proved that the agent choose to execute the ``least
unethical'' plan it had available.

\section{Further Work}
We believe that there is a value in the existence of a declarative
language for describing consequence engines and that the \ail-based
implementation used in this verification lays the groundwork for such
a language.  We would be interested in combining this language, which
is structured towards the ethical evaluation of actions, with a
language geared towards the ethical evaluation of plans for BDI
systems, such as is discussed in~\cite{dennis14ethical}.

While using \ajpf\ allowed us to very rapidly implement and verify a consequence engine in a scenario broadly similar to that reported in Winfield et al.~\cite{Winfield13} there were obvious issues trying to adapt an approach intended for use with BDI agent languages to this new setting.

In order to verify the consequence engine in a more general, abstract,
scenario we had to endow it with mental states and it may be
appropriate to pursue this direction in order to move our declarative
consequence engine language into the sphere of BDI languages.  An
alternative would have been to equip \ajpf\ with a richer property
specification language able to detect features of interest in the
ethical selection of actions.  At present it is unclear what such an
extended property specification language should include, but it is
likely that as the work on extending the declarative consequence
engine language progresses the nature of the declarative properties to
be checked will become clearer.  It may be that ultimately we will
need to add BDI-like features to the declarative consequence engine
\emph{and} extend the property specification language.

We would also like to incorporate the experimental validation approach
into our system by using the \mcapl{} framework's ability to integrate
with the Robot Operating System~\cite{ros-ail-TR} in order to use our
new ethical consequence engine to govern actual physical robots in
order to explore how formal verification and experimental validation
can complement each other.

\section{Conclusion}

In this paper we have constructed an executable model of an ethical
consequence engine described in~\cite{Winfield13} and then verified
that this model embodies the ethical principles we expect.  Namely
that it pro-actively selects actions which will keep humans out of
harms way, if it can do so.  In the course of developing this model we have
laid the foundation for a declarative language for expressing ethical
consequence engines.  This language is executable and exists within a
framework that can interface with a number of external robotic systems
while allowing elements within the framework to be verified by model
checking.  

At present the language is very simple relying on prioritisation first over individuals and then over outcomes.  It can not, for instance, express that while, in general, outcomes for individuals of some type (e.g., humans) are more important than those for another (e.g., the robot) there may be some particularly bad outcomes for the robot that should be prioritised over less severe outcomes for the humans (for instance it may be acceptable for a robot to move ``too close'' to a human if that prevents the robot's own destruction).  Nor, at present, does the language have any ability to distinguish between different \emph{contexts} and so an outcome is judged equally bad no matter what the circumstances.  This will be too simple for many situations --  especially those involving the competing requirements of privacy and reporting that arise in many scenarios involving robots in the home.  The language is also tied to the existence of an engine that is capable of simulating the outcomes of events and so the performance of a system involving such a consequence engine is necessarily limited by the capabilities of such a simulator.  This simulation is tied to a single robot action and so, again, the system has no capability for reasoning that some action may lead it into a situation where the only available subsequent actions are unethical.  Lastly the language presumes that suitable ethical priorities have already been externally decided and has no capability for determining ethical actions by reasoning from first principles.

Nevertheless we believe that the work reported here opens the path to a system for implementing verifiable
ethical consequence engines which may be interfaced to arbitrary
robotic systems.

\section{Software Archiving}
The system described in this paper is available as a recomputable virtual machine on request from the first author and will be archived at \url{recomputation.org} in due course.  It can also be found on branch \texttt{ethical\_governor} of the git repository at \url{mcapl.sourceforge.net}.  

\section{Acknowledgements}

Work funded by EPSRC Grants EP/L024845/1 and EP/L024861/1
(``Verifiable Autonomy'').

\bibliographystyle{abbrv}
\bibliography{references}
\end{document}